\documentclass[lettersize,journal]{IEEEtran}
\usepackage{amsmath,amsfonts}
\usepackage{algorithmic}
\usepackage{algorithm}
\usepackage{array}
\usepackage[caption=false,font=footnotesize,labelfont=rm,textfont=rm]{subfig}
\usepackage{textcomp}
\usepackage{stfloats}
\usepackage{url}
\usepackage{verbatim}
\usepackage{graphicx}
\usepackage{cite}
\usepackage{amssymb}
\usepackage{multirow}
\usepackage{graphicx}

\begin{document}
	\title{Style Alignment based Dynamic Observation Method for UAV-View Geo-localization}
    
	\author{Jie Shao,  \emph{Member, IEEE}, LingHao Jiang
    \thanks{Jie Shao (e-mail:shaojie@shiep.edu.cn) and LingHao Jiang (e-mail:linghaojiang@mail.shiep.edu.cn) are with the Department of Electronics and information engineering, Shanghai University of Electric Power.}
	}
   
	\maketitle
    
	\begin{abstract}
		The task of UAV-view geo-localization is to estimate the localization of a query satellite/drone image by matching it against a reference dataset consisting of drone/satellite images. Though tremendous strides have been made in feature alignment between satellite and drone views, vast differences in both inter and intra-class due to changes in viewpoint, altitude, and lighting remain a huge challenge. In this paper, a style alignment based dynamic observation method for UAV-view geo-localization is proposed to meet the above challenges from two perspectives: visual style transformation and surrounding noise control. Specifically, we introduce a style alignment strategy to transfrom the diverse visual style of drone-view images into a unified satellite images visual style. Then a dynamic observation module is designed to evaluate the spatial distribution of images by mimicking human observation habits. It is featured by the hierarchical attention block (HAB) with a dual-square-ring stream structure, to reduce surrounding noise and geographical deformation. In addition, we propose a deconstruction loss to push away features of different geo-tags and squeeze knowledge from unmatched images by correlation calculation. The experimental results demonstrate the state-of-the-art performance of our model on benchmarked datasets. In particular, when compared to the prior art on University-1652, our results surpass the best of them (FSRA), while only requiring 2× fewer parameters. Code will be released at https://github.com/Xcco1/SA\_DOM
	\end{abstract}
	
	
	\section{Introduction}
	\IEEEPARstart{C}{ross}-view geo-localization is the task that matches the geographic images from different views, such as homologous architecture localization among ground-view, drone-view, and satellite-view images\cite{2020Where,liu2019lending,9684950}. Due to the popularity of drones, UAV-view geo-localization related applications between drone-view and satellite-view images attract a lot of attention, such as autonomous vehicles, drone navigation, event detection, accurate delivery, and so on\cite{2022arXiv220701768W,2022arXiv220611499J}.

    Similar intra-class and inter-class distances are the main challenges of UAV-view geo-localization. Specifically, satellite images and drone images have different imaging methods, perspectives, and fields of view, resulting in significant intra-class differences. For instance, satellite images are generated by a ranging (LiDAR) sensor, which has a large field of view, showing only the top of architectures. Drone images are from an RGB camera with varied lighting conditions, which has a smaller Fov and shows one side of the architecture. Meanwhile, most architectures/locations share similar building styles and homogeneous appearances, which lead to smaller inter-class differences.
    
    Most existing progress on this topic focuses on extracting salient features to expand inter-class distances, including keypoint matching and partition alignment. Keypoint matching may fail if there are not many corresponding local features between the two view pairs, so it is mostly used as an auxiliary method recently\cite{9779991}. Square-ring based partitioning is the most widely used feature partition alignment method, which usually splits the images into four square-ring equal-importance parts and concatenates them to one, like Local Pattern Network (LPN)\cite{wang2021each}, PCL\cite{tian2021uav}, etc. Such approaches have the advantage of overcoming different rotational angles of drones and are easy to implement. But four equal-importance partitioning have the potential to introduce surrounding noise caused by various flight altitudes. For example, different altitudes make different image contents and distributions, so a fixed image partition will result in different contents in the patches at the same position. Besides, surrounding noise will be inevitably increased if we treat all the partitions the same. 

    To alleviate surrounding noise and expand inter-class distances, human observation habit has presented a cue. We notice that images of the same geo-tag will share a common geographic coordinate point as the center, and there are usually one satellite image and dozens of drone images with different angles and altitudes taken around that coordinate point. Under this prior knowledge, humans tend to first find whether there is a salient target in the central part of the image and use it as a reference point for image matching, while the surroundings other than the salient target are only used as auxiliary information. But if there is no salient object in the center, humans would search for more useful information in the surroundings. Therefore, we simulate human observation by dividing the image into the center and the surroundings, and dynamically generate feature representations for image matching based on the presence of salient targets in the central region. 
    
    On the other hand, intra-class distance narrowing has not been well explored. we notice that satellite images usually have a homogeneous color style due to their image generation method. In contrast, the visual styles of drone images vary with changes in light intensities and camera parameters. The diverse visual representation of the drone images and their differences from the satellite images increase the intra-class distances. Therefore, it would be a great help to image matching by transferring the visual style of the satellite image to the drone image so that they both have a stable and similar color style. However, although there are many deep learning methods for image style transfer like CycleGAN\cite{Zhu2017UnpairedIT}, Pix2Pix\cite{Isola2016ImagetoImageTW}, etc., they all have difficulty in accurately separating the image style from the image content, even resulting in semantic features of the image being ruined. Therefore, we design a style alignment strategy based on the traditional image processing method to achieve the goal.
    
	Based on the above considerations, we propose a style alignment based dynamic observation method for UAV-view geo-localization. Specifically, we transfer the visual styles of drone images to satellite before feature extraction first, and then leverage a dynamic observation module to generate feature representation in a human observation manner. Later, features are put into the classifier for geo-tag labeling. Different from other approaches using cross-entropy loss, a deconstruction loss is proposed to squeeze out the correlation knowledge from the negative predictions for feature representation training.
	
	In summary, the main contributions of this paper are as follows:
	\begin{itemize} 
		\item  We propose a style alignment strategy to effectively solve the problem of intra-class distances due to visual style discrepancies (caused by lighting conditions and camera parameters). As far as we know, we are the first to bridge the view gap in cross-view geo-localization by style alignment. 
		\item  We propose a dynamic observation module that solves the problem of surrounding noise by flexibly generating feature representation based on the position of the salient target, imitating human observation habits
		\item  A loss combination is proposed by introducing a deconstruction loss to achieve not only an improvement in the similarity of samples of the same class during model training, but also a reduction in the correlation of non-similar features based on the negative part of the predicted probability distribution.
		\item  We not only achieve state-of-the-art performance on the task but also limit our parameter size to half of FSRA\cite{2022A}, who published the SOTA results. Besides, our experimental results on transfer learning from University-1652 to Sues-200\cite{2022arXiv220410704Z} also demonstrate robust performance.	
	\end{itemize}
	\section{Related work}
	\subsection{UAV-view geo-localization}
	With the development of drone, the applications based on drone-view and satellite-view image matching is increasing, such as aerial photography\cite{2022arXiv220611499J}, target tracking\cite{2022arXiv220701768W}, drone navigation\cite{2022Vision}, scene classification\cite{9598903}, accurate delivery, and so on. Accordingly, many works are devoted to drone and satellite-view geo-localization.
	
	Current works of UAV-view geo-localization were all based on the dataset University-1652\cite{zheng2020university} proposed by Zheng, who tried two missions including drone view target localization and drone navigation. He introduced a simple baseline as well. Contextual correlation and feature alignment are two main challenges in the task. Wang et al.\cite{wang2021each} proposed the famous square-ring partition to consider surrounding information and confront rotation, then Dai et al.\cite{2022A} proposed a transformer baseline and automatically divided images into different semantic parts with heatmaps. Inspired by the above two works and lots of works in person Reid\cite{2017Beyond,2017arXiv171108184Z}, we noticed that partition-based feature alignment is useful in taking full advantage of surrounding information. Besides, the domain gap between the drone view and satellite view is also considerable, so Tian et al.\cite{tian2021uav} transformed the drone view image into the satellite view to confront the different view angles. Moreover, Wang et al.\cite{2022Multiple} proposed a network to dynamically adjust the domain shift caused by environmental variation. The imbalance of training data was also concerned by researchers. Ding et al.\cite{ding2021practical} proposed data enhancement on satellite view images to solve the lack of satellite images. To explore the effect of different flight altitudes, recently Zhu.\cite{2022arXiv220410704Z} proposed a new multi-height, multi-scene dataset SUES-200, which provides drone images at different flight altitudes.
	
	\subsection{Loss functions in cross-view geo-localization}
	Suitable and efficient loss functions are important for model training and feature extraction. Nowadays, the Cross-Entropy loss\cite{2017A}, Triplet loss\cite{2015arXiv150303832S}, and Contrastive loss\cite{2006Dimensionality} are widely deployed in the task of image retrieval and re-identification\cite{2022Hierarchical}. In cross-view geo-localization, Zheng et al.\cite{zheng2020university} applied the Instance loss\cite{2020Dual} to treat each sample as a class and used the verification loss to measure the relationship between samples. Thereafter, Hu et al.\cite{hu2018cvm} proposed a weighted soft margin ranking loss to speed up the training convergence and improve the retrieval accuracy. Considering the different feature spaces between the Triplet loss and the ID loss, Luo et al.\cite{2020A} proposed a BNNeck to put features into the corresponding feature space of ID loss and Triplet loss. Furthermore, to narrow the intra-class distance of the same geo-tag, Dai et al.\cite{2022A} introduced the KL loss to gather similar instances in drone-view and satellite-view images. However, these losses usually implement on the features. Especially the cross-entropy loss, which is used by most of the cross-view geo-localization methods. But the cross-entropy only considers the predicted probability of the class that matched the true labels of the current example, while ignoring information on the negative predicted distribution. So we introduce the deconstruction loss as a compliment for the cross-entropy to gather features of the same geo-tag and reduce correlations among different geo-tag features during training.
 
    \subsection{Attention mechanism}
    It is well known that attention plays an important role in human perception. One important property of a human visual system is that human beings do not capture everything equally at once. Instead, he/she exploits a sequence of partial glimpses and selectively focuses on salient parts to build the visual structure. Hu et al.\cite{hu2018squeeze} introduced a compact module to adaptively recalibrate channel-wise features by explicitly modeling interdependencies between channels. To fuse spatial attention, Woo et al.\cite{woo2018cbam} proposed a convolutional block attention module along two separate dimensions, channels, and spaces. Moreover, attention mechanisms are widely used in the field of remote sensing. Remote sensing data often exhibit rich semantic spatial relationships between pixels, along with specific underlying associations, such as the combination of urban high buildings and green landscapes, or the agricultural land and lakelet. Pan et al.\cite{rs11080917} introduced the spatial and channel attention mechanisms to selectively enhance more useful features in specific positions and channels, further, Hong et al.\cite{GraphCN} proposed a mini-GCNs branch to model middle-range and long-range spatial relations between samples through their graph structure in UAV view. It is worth noting that most attention methods aimed to design a general attention block to capture long-range dependencies. However, in specific downstream tasks like UAV-view geo-localization, exploiting additional spatial distribution information would reduce the complexity of the attention mechanism and improve its performance. To this end, inspired by the human observation habit, which assumes that the surroundings are unnecessary when the center part has well-marked features, we model the dynamic relationship between the surroundings and the center by the hierarchical attention block (HAB).
	\section{Proposed method}
        \begin{figure*}
		\centering
		\includegraphics[width=0.9\linewidth]{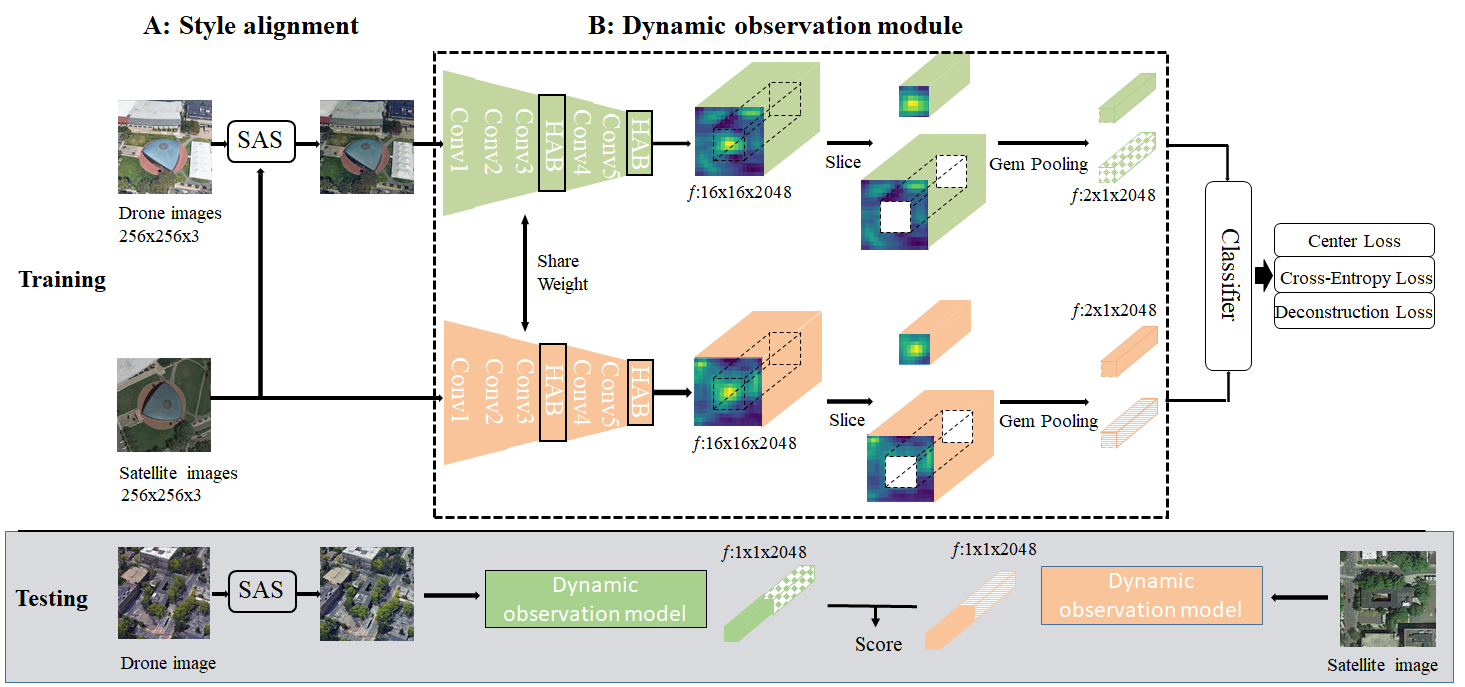}
		\captionsetup{justification=centering}
		\caption{The framework of the proposed method. It is composed of part A, part B, and the classifier. Part A is the preprocessing stage which aligns the visual styles of the two view images by SAS. Then satellite-view images and preprocessed drone-view images are put into part B with two separate network branches. The two branches are of the same structure and share weights. }
		\label{Fig.1.}
	\end{figure*}
	In this section, the structure of our style alignment based dynamic observation method is introduced in Fig. 1. Firstly, the style alignment strategy is applied to the drone-view images in part A. Next, part B is the dynamic observation module, which is designed to extract features imitating a human being. It includes the ResNet-GL associated with the dual square-ring strategy and the Gem Pooling layer. In ResNet-GL, the hierarchical attention blocks (HAB) are proposed to decide if there is a salient geographic object in the center. Finally, the loss combination supervises the learning of feature representations of every geo-tag.

	\subsection{Style Alignment Strategy (SAS)}
\begin{figure*}[htb]
		\centering
		\includegraphics[scale=0.7]{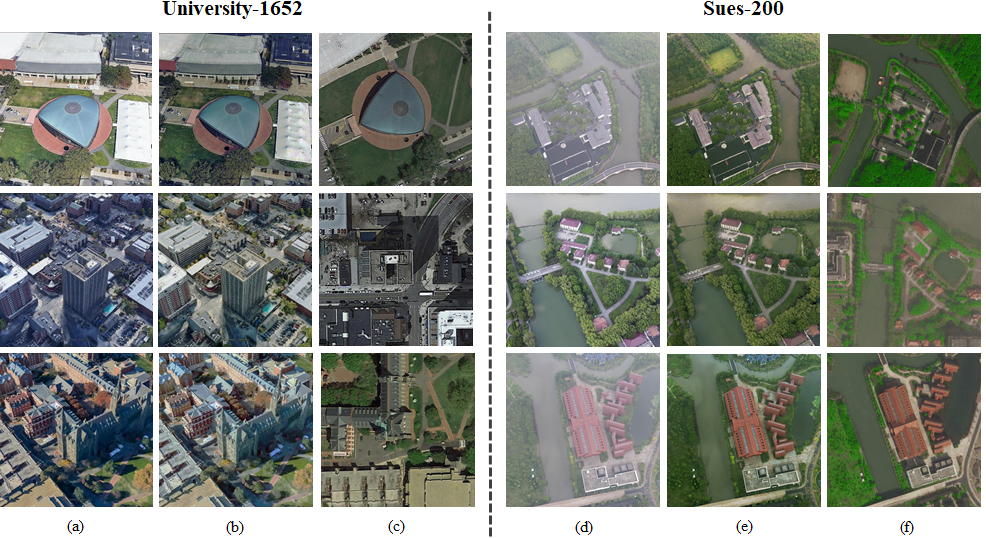}
		\captionsetup{justification=centering}
		\caption{Examples of the SAS results. The results of University-1652 are on the left side and the other is the results of Sues-200. (a) and (d) are the original drone-view images, (b) and (e) are results of the SAS, whose visual styles are similar to the satellite images, and (c) and (f) are the satellite view images.}
		\label{Fig.2.}
	\end{figure*}

	Intra-class distances between a pair of drone-view and satellite-view images include not only disparity of view angles but also varieties of visual styles. As we can see in Fig. \ref{Fig.2.} (c) and (f), the satellite-view images usually have a similar visual style, but the drone-view images in Fig. \ref{Fig.2.} (a) and (d) are variable due to different views, light conditions, and camera chromatic aberrations. Therefore, style alignment could assist feature alignment to narrow intra-class distance. There are three types of style transfer methods currently available: generative adversarial network-based methods, image-adaption-based methods, and traditional image processing methods. Our goal is to achieve style transfer without changing the texture and semantic content of the image. Additionally, we need to generate models based on a small number of samples (limited by the datasets) and preferably with low system requirements. Finally, through experimental analysis and reference to CDNet\cite{9386248}, we design our style transfer algorithm based on the nonlearning method. The discussions can be found in Section \uppercase\expandafter{\romannumeral4}.\emph{F}.
 
    Our style alignment strategy (SAS) makes the drone images imitate satellite images in visual style at the stage of preprocessing. Specifically, the SAS needs no training, so it does not increase the training burden. The implementation details are as follows. Firstly, the cumulative distribution in RGB channels of the satellite image is calculated to get the transformation mapping function $M(x)$, as shown in Equ. 1:
	$$
	\begin{aligned}
		&E_{c}(x)=H_{c}(x) / \sum_{x=0}^{255} H_{c}(x) \\
		&P_{c}(x)=(\sum_{i=0}^{x} E_{c}(i)) * 255+0.5 \\
		&\operatorname{M}(x)=\left[P_{R}(x), P_{G}(x), P_{B}(x)\right] \\
	\end{aligned}
	\eqno{(1)}	
	$$
	where $E_{c}(x)$ is the normalized number of pixels with value $x$ ($x \in [0, 255]$) in $R/G/B$ channel of the satellite image. Then we calculate the cumulative distribution of $E_{c}(x)$ and extend their value to the range of 0 to 255 to get $P_{c}(x)$. $M(x)$ is assumed to be the visual style features of the satellite image through the gray-level statistical distribution of RGB channels, where $R/G/B$ in $P_{R/G/B}(x)$ stands for $R/G/B$ channel respectively.
	
	Considering a single image could not represent all the satellite images, we calculate the whole training set to get the average mapping function $\hat{M}(x)$, as shown in Equ. 2, where $M(x_{k})$ is the transformation map generated by the k-th satellite images, and $N$ is the number of satellite images in the training set.	
	$$
	\operatorname{\hat{M}}(x)=\left(\sum_{k=1}^{N} \operatorname{M}\left(x_{k}\right)\right)/N
	\eqno{(2)}	 
	$$
	\begin{figure}[htb]
		\centering
		\includegraphics[width=1\linewidth]{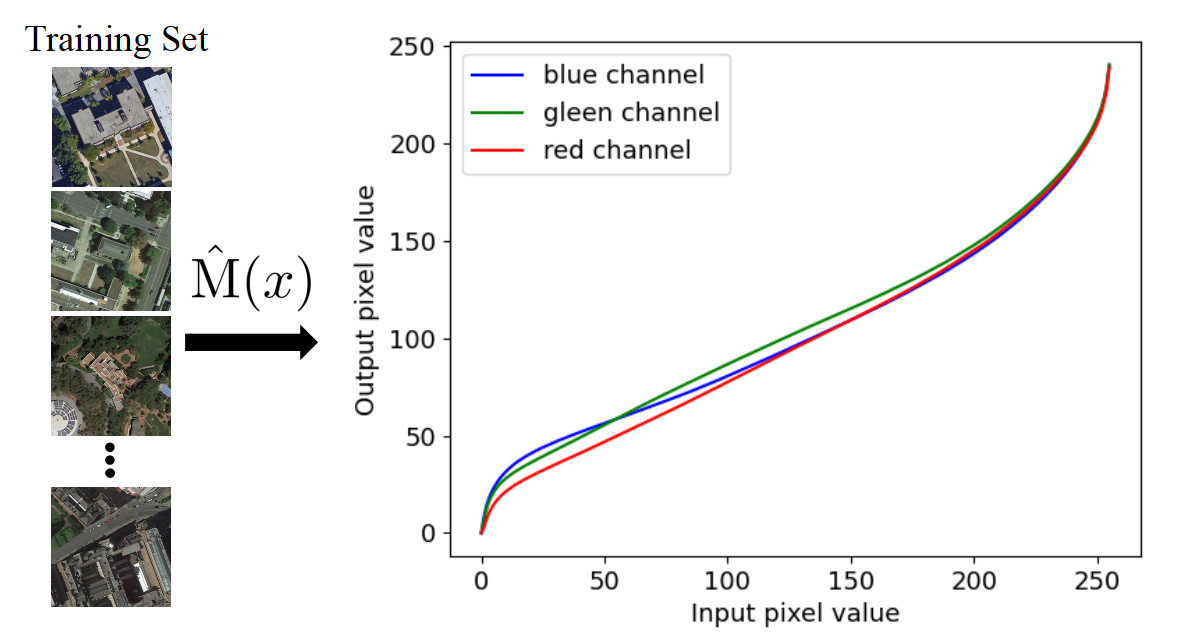}
		\captionsetup{justification=centering}
		\caption{The transformation map of the whole train datasets, three curves represent the RGB channels respectively.}
		\label{Fig.3.}
	\end{figure}
	
	The transformation map of the satellite set is illustrated in Fig. \ref{Fig.3.}, where the x-axis indicates the variation of input pixel value from 0 to 255, and the y-axis indicates the output of SAS. It can be seen that at the beginning stage, all three curves have a sudden rise, which could fix the shadow and the shaded side of the building in the drone-view images, as shown in the second and the third row in Fig. \ref{Fig.2.}(a). Thereafter, the green curve is slightly higher than the others in the middle of the map, and the red curve is always relatively lower than the others. They reflect the retro looks of the satellite view.

	Then the transformation process of the drone images is shown in Equ. 3:
	$$
	I^{'}=\{\operatorname{\hat{M}}(z)\}=\{\left[\operatorname{\hat{P}}_{R}\left(z_{R}\right), \operatorname{\hat{P}}_{G}\left(z_{G}\right), \operatorname{\hat{P}}_{B}\left(z_{B}\right)\right]\}
	\eqno{(3)}	
	$$
	$ z_{R,G,B}$ indicates the gray level of the each pixel $z$ in RGB channel of the input drone image $I$, and $ I^{'}$ is the mapping output image. Thus the drone images are transferred to the satellite visual style. 
	
	Fig. \ref{Fig.2.} shows the aligned results of drone-view images of University-1652 in (b). The original drone-view images are in strong sunlight or backlight. It can be seen that their visual styles are close to the satellite images after running SAS. The results of Sues-200 are shown in Fig. \ref{Fig.2.} (e). The original images with different lighting conditions are unified into one visual style. As a result, not only the view gap is reduced, the feature discrepancies caused by the lights are reduced as well. The SAS is demonstrated to be robust on different datasets as well.
	
	\begin{figure}[htb]
		\centering
		\includegraphics[width=0.8\linewidth]{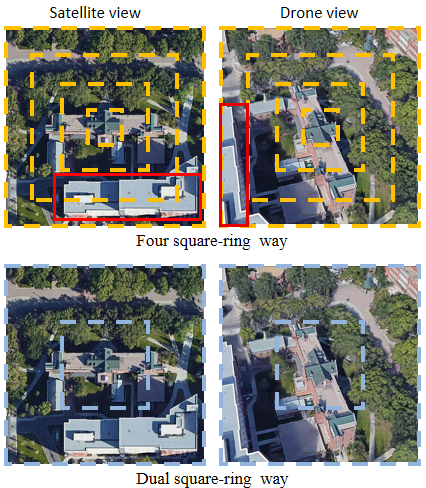}
		\caption{Illustration of the four square-ring partition (top) and our dual square-ring partition (bottom) by an example pair of images. In both two cases, there are a pair of the satellite-view image and the drone-view image shown in the first and the second column respectively. In the top row, the architecture in the red box is partitioned into two parts in the satellite view, but in the drone view, it is in one part, which would cause mismatching by the model.}
		\label{fig.4.}
	\end{figure}
	\subsection{Dynamic Observation Module}
	To expand the inter-class distance and decrease surrounding noise, we propose a dynamic observation module by flexibly generating feature representation based on the position of the salient target, imitating human observation habits. It follows the style alignment strategy. It mainly includes the ResNet-GL, the dual square-ring partition strategy, and the generalized-mean pooling part.

	\textbf{The dual square-ring strategy:} Four-part square-ring partition is currently a widely used feature partition method for cross geo-localization. Since satellite and UAV images of the same geo-tag are taken at the same geographic coordinates, segmenting the image into ring regions for feature characterization can obtain more reasonable image semantics than other partition ways. On the first row of Fig. 4 are segmentation examples of the four-part square-ring partition method, and the bottom rows are the dual-part square-ring method. As the four-part square-ring partition method divides the image into four parts, each region is relatively small, and geographic landmarks (buildings, squares, etc.) are often divided into different patches. This leads to two problems: 1. The semantics of the same geographic landmark are destroyed when they are separated into different partitions; 2. The calculation and weight of each partition are the same, so the influence of surrounding noise on feature representation is relatively large.
 
    Based on the above considerations, our dual square-ring partition treats the image from three aspects: the center, the surroundings, and the whole, as shown in Fig. 4 (bottom). On one hand, segmenting the image into two patches will greatly reduce the probability of the geographic landmarks being segmented into different patches; on the other hand, combined with HAB, different feature extraction methods will be applied to the center and surrounding, thus achieving dynamic representation of the image features.
	    \begin{figure*}[htb]
		\centering
		\includegraphics[width=1\linewidth]{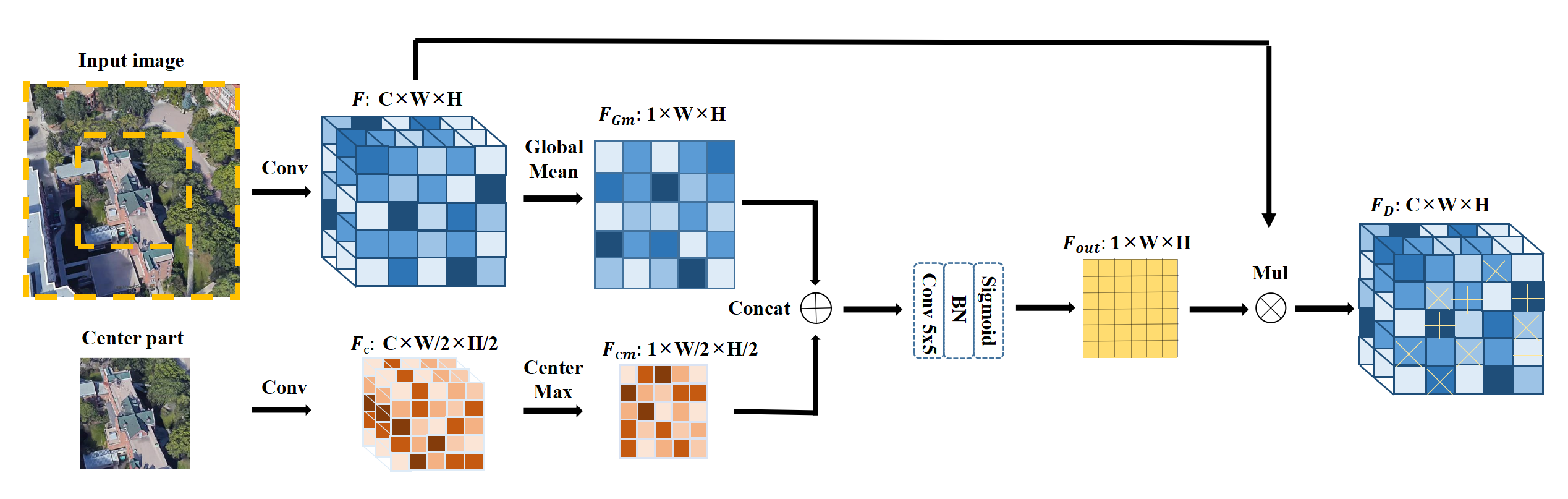}
		\caption{The architecture of the hierarchical attention block (HAB), which is fabricated in a dual-square-ring stream network.}
		\label{fig.5.}
	\end{figure*}

	\textbf{The hierarchical attention block:} The most important part of the ResNet-GL structure is the  Hierarchical Attention Block (HAB). It is designed based on the idea that humans usually look for a geographic marker at the center of the image and use it as the main basis for image matching if it exists, or match based on the whole image if it is not. Combined with the dual square-ring strategy, we divide the image into two branches of the center part and the whole part using different methods for attentive calculation, as shown in Fig. 5. The branch of the center part is used to obtain information on salient geographic markers, and the branch of the whole part is used to extract surrounding information prepared for subsequent convolution fusion. A convolution fusion block is used to determine whether the center part is discriminative enough or how much information should be taken from the surroundings. The whole image instead of the surroundings is applied in HAB to reduce possible semantic misalignment caused by the dual square-ring partition strategy. 
 
	To this end, if a feature map $ F\in \mathbb{R}^{C\times W\times H} $ is given, they will be put into two branches first: one implements the max operation along the channel axis on the center part of $ F $ and the other implements the mean operation along the channel axis for every spatial index, where the center part is defined by our two-part square-ring partition strategy. Then the two attention maps $ F_{Cm}\in \mathbb{R}^{1\times W/2\times H/2} $ and $ F_{Gm}\in \mathbb{R}^{1\times W\times H} $ are obtained from the two branches, where the subscript $Cm$ and $Gm$ denote the center part max operation and the global part mean operation respectively. To align the sizes of $ F_{Cm} $ and $ F_{Gm}$, we pad zeros around $ F_{Cm}$ to make it the same size as $F_{Gm}$. Later they are concatenated and processed through the $ 5\times 5$ convolution layer ($Conv_{5 \times 5}$), the batch normalization (BN), and the sigmoid operations to handle two branches of information. Consequently, the prominent feature of the center architecture could be extracted by the first branch, while the second branch keeps the surrounding information if the center part does not have prominent features. The final output of HAB is $ F_{out}\in \mathbb{R}^{1\times W\times H} $. We multiply the input $ F $ and the output $ F_{out} $ to get the dynamic-aware feature $ F_{D} $. The above procedure is formulated in Equ. 4.
	$$
	\begin{gathered}
		F_{\text {cat }}=\text { Concat }(\text { Gm }(F), \text { Cm }(F)) \\
		F_{\text {out }}=\text { Sigmoid }\left(\operatorname{BN}\left(\operatorname{Conv}_{5 \times 5}\left(F_{\text {cat }}\right)\right)\right) \\
		F_{D}=F * F_{out}
	\end{gathered}
	\eqno{(4)}	 
	$$
	\textbf{Generalized-mean Pooling:} Significant information about the surroundings would be diluted by noises if the global average pooling is applied on the feature map. As a result, we introduce the Generalized-mean Pooling (Gem pooling)\cite{8382272} here. It has an excellent performance in various retrieval tasks, such as Person ReID\cite{arxiv20reidsurvey}, Instance retrieval\cite{deng2022insclr}, etc. In our approach, Gem pooling can be more effective in preserving surrounding information in a large size feature map. The output feature map of ResNet-GL is a tensor of $2048\times 16\times 16$, where the center part is only represented by the size of  $2048\times 8\times 8$ and the rest features represent the surroundings, which are three times the size of the center features. The calculation of Gem pooling is shown in Equ. 5.
	$$
    f_{Gem}=\left[f_{1} \ldots f_{k} \ldots f_{C}\right]^{\top},f_{k}=\left(\frac{1}{\left|F_{k}\right|} \sum_{\theta \in F_{k} }\theta^{p}\right)^{\frac{1}{p}}
	\eqno{(5)}	
       $$
    \noindent where $ f_{k}, k\in(1,2,..,C)$ represents the $k$th channel of Gem pooling result. $F_{k}$ is the $k$th channel of the ResNet-GL result, $ F_{k}\in \mathbb{R}^{1\times W\times H}$. The output of the Gem pooling, $f_{Gem} \in (C,1,1)$, consists of a single value per feature map. $p$ is the hyper-parameter of pooling. The global max pooling and global average pooling are two special cases of Gem pooling, i.e., max pooling when $p$→$ +\infty$  and average pooling for $ p$ =1. We set the default value to 3 in our experiments. 

	\begin{figure}[t!]
		\centering
		\includegraphics[width=1\linewidth]{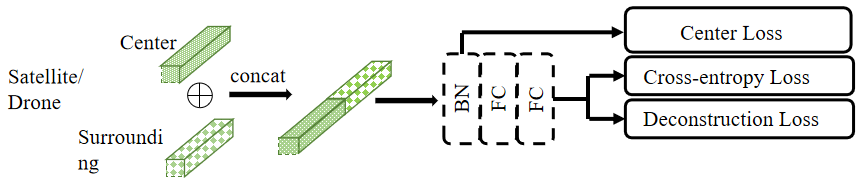}
		\caption{The structure of the classification module. The green bars are the features of the drone view or the satellite view output from the Gem pooling layer. One of them is the feature map of the image center, and the other is that of the surroundings. The two parts are concatenated and put into a BatchNorm layer, followed by two fully connected layers to compress the feature dim and predict the geo-tags respectively. The Center loss, the Cross-entropy loss, and the Deconstruction loss are calculated finally.}
		\label{fig.6.}
	\end{figure}
	\subsection{The Loss Design}
	After feature extraction, the classification module is employed to supervise learning on geo-tag features. The triplet loss and cross-entropy loss are the most common loss combinations applied in cross-view geo-localization works\cite{zheng2020university,zhuang2021faster,wang2021each,tian2021uav}, where the triplet loss is used to narrow the distance of the same geo-tag in each view, and the cross-entropy loss is for classification. Inspired by the idea of the re-id problem, we introduce the center loss\cite{2016A} instead of the triplet loss, as the former could learn a center for each class, so that the feature distribution of each class would be more concentrated, as formulated in Equ. 6.
        $$
	\mathcal{L}_{Center}=\frac{\gamma}{2}\sum_{n}\left(\vert\vert x_{n}-c_{n} \vert\vert  \right)^{2}
	\eqno{(6)}
	$$
    $c_{n}$ denotes the $n$th class center of features, and $x_{n}$ denotes each feature of the $n$th class. In practice, we also need to add a parameter $\gamma$ to balance the weights before other losses. 
    
    Besides, the cross-entropy loss only cares about the predicted probability of the class that matched the true labels, but discards the information of negative samples. As formulated in Equ. 7, $y_{n}$ denotes $n$th class in the one-hot vector generated by the ground-truth $y$, and $\hat{y}$ is the predicted vector from the classifier.
        $$
	\mathcal{L}_{CE}(y,\hat{y})=- \sum_{n} y_{n} \log(\hat{y}_{n})
	\eqno{(7)}
	$$ 
    Therefore, we propose a deconstruction loss (Dc loss) as a complement to the cross-entropy loss, which will push away different geo-tag features by correlation calculation on negative predictions. The details of the classification module are shown in Fig. 6. 

    The feature maps of drone-view and satellite-view images are concatenated and normalized, and then sent to calculate the center loss. Later, two fully connected layers are applied to compress the feature dim from 4096 to 2048 and predict the geo-tags respectively. Then, the cross-entropy loss and deconstruction loss are applied. We first calculate the cross-correlation matrix of the predicted probability, then reduce the cross-correlation of different geographical locations to zero and push self-correlation to one, as shown in Equ. 8. To calculate the cross-correlation matrix, we take the predicted probability vectors for each example and stack them together into a matrix of size N x C. Then compute the correlation matrix for this N x C matrix, which is a C x C matrix. Each item represents the correlation between the predicted probabilities of two different classes. The purpose is that geographical locations are independent, so the cross-correlation of different locations should be close to zero.
        
	$$
	\mathcal{L}_{D c}=\sum_{i}\left(1-S_{i i}\right)^{2}+\lambda \sum_{i} \sum_{j \neq i} S_{i j}^{2}
	\eqno{(8)}
	$$
	where $S$ is the cross-correlation matrix of $\hat{y}(x)$, which is a matrix of geo-tag numbers by geo-tag numbers. $\hat{y}(x)$ is the predicted probability of the classifier. $S_{i i}$ is the diagonal element of $S$ and  $S_{i j}$ is the off-diagonal element of $S$. $\lambda$ is used to control the contribution of the off-diagonal elements, which is set to 0.2. 
	
	\section{Experiments}
	In this section, we first evaluate our approach and compare it with some state-of-the-art methods on datasets of University-1652\cite{zheng2020university} and Sues-200\cite{2022arXiv220410704Z}. Then ablation studies are performed on important parts of our approach. Later, the results of transfer learning are shown to verify the robustness of our method as well. Discussions and visualizations are listed at the end.
	
	\subsection{Datasets and Evaluation Strategy}
	{\bf Datasets}: Two benchmark drone-to-satellite image geo-localization datasets: University-1652 and Sues-200 are applied in our experiments. the University-1652 dataset contains 1652 buildings from 72 universities around the world. It provides images of three views for each building, including one satellite view image, 54 drone view images with different flight heights and angles, and several ground-view images. We only use satellite images and drone images for UAV-view geo-localization.
	
	The Sues-200 dataset was put forward recently, which contains 200 locations in the vicinity of SUES (Shanghai University of Engineering Science). It provides 50 drone-view images and one satellite-view image at one location. Different from University-1652, it divides the drone view images with different flight altitudes at 150m, 200m, 250m, and 300m. 
	
	As the Sues-200 is a new dataset, which does not have many experimental results, we compare our results with other methods and the ablation study on the University-1652. Both two datasets are utilized for experiments on transfer learning. To make a fair and reliable comparison, we use the consistent training and testing set in this study.
	
	{\bf Evaluation Protocols}: Same with most of the recent works, we use Recall@K and Average Precision (AP) to evaluate the performance of our model, which are both widely used in image retrieval. Recall@K represents the probability of the correct matching in the top K of the retrieval rankings, so it is sensitive to the position of the first true matching image. AP considers the position of all the true matching images in the evaluation, which is the measurement of the overall performance of the model.
	
	\subsection{Implementation Details}
	{\bf In network structure and training strategy}. The ResNet50 is applied as the backbone, but the stride of its second convolutional layer is modified and the last downsampling layer of conv5 is also reduced from 2 to 1. Then the pre-trained ImageNet ResNet50 model is loaded. As the drone and satellite-view images are separately put into two parallel network branches, their weights are updated and shared after every epoch. Input images are set to 256 × 256 pixels both in training and testing. To save memory space and keep the training in an end-to-end way, the style alignment strategy is implemented before data augmentation.
	
	During training, we use the stochastic gradient descent (SGD) with a momentum of 0.9 as the optimizer and a weight decay of 0.0005 with a mini-batch of 32. The initial learning rate of the backbone layer is 0.01, then it decreased by 0.5 after 30 epochs. The whole training includes 200 epochs. Our model is implemented on Pytorch, and all experiments are conducted on one NVIDIA RTX 1080Ti GPU. During testing, we utilize the Euclidean distance to measure the similarity between the query image and candidate images in the gallery.
	
	{\bf In the loss function}. The center loss is used to narrow the distance of the same geo-tag in each domain, and the cross-entropy loss combined with the deconstruction loss is applied for classification. Besides, the weights of the Center loss, the Cross-entropy loss, and the Deconstruct Loss are set to 0.0005:1:1 in experiments. 
	
	\begin{table}[htb]
		\centering
		\caption{comparison with state-of-the-art methods on University-1652. The best results are shown in {\bf BLOD} }
		\label{Tabel1}
		\renewcommand\arraystretch{1.5}
		\begin{tabular}{c|cccc}
			\hline
			\multirow{3}{*}{Method} & \multicolumn{4}{c}{University-1652}                                                                      \\ \cline{2-5} 
			& \multicolumn{2}{c|}{Drone -\textgreater Satellite}   & \multicolumn{2}{c}{Satellite -\textgreater Drone} \\
			& R@1            & \multicolumn{1}{c|}{AP}             & R@1                     & AP                      \\ \hline
			Zheng et al.\cite{zheng2020university}            & 59.69          & \multicolumn{1}{c|}{64.8}           & 73.18                   & 59.4                    \\
			LCM\cite{ding2021practical}                     & 66.65          & \multicolumn{1}{c|}{70.82}          & 79.89                   & 65.38                   \\
			LPN(w/o Google)\cite{wang2021each}         & 74.18          & \multicolumn{1}{c|}{77.39}          & 85.16                   & 73.68                   \\
			LPN\cite{wang2021each}                     & 75.93          & \multicolumn{1}{c|}{79.14}          & 86.45                   & 74.79                   \\
			LPN+CA-HRS \cite{lu2022content}                     & 76.67          & \multicolumn{1}{c|}{79.77 }          & 86.88                   & 74.83                 \\
			LPN+USAM\cite{9779991}                & 77.60          & \multicolumn{1}{c|}{80.55}          & 86.59                   & 75.96                 \\
			PCL\cite{tian2021uav}                     & 79.47          & \multicolumn{1}{c|}{83.63}          & 87.69                   & 78.51                   \\
			FSRA(k=1)\cite{2022A}               & 82.25          & \multicolumn{1}{c|}{84.82}          & 87.87                   & 81.53                   \\ \hline
			\textbf{Ours}           & \textbf{84.08} & \multicolumn{1}{c|}{\textbf{86.39}} & \textbf{91.44}          & \textbf{82.02}           \\ \hline
		\end{tabular}
	\end{table}
	
	\subsection{Comparison with Other Methods }
	We compare our method with existing competitive methods on the University-1652 dataset with an image size of 256 × 256px in Table \ref{Tabel1}. For a fair comparison, their reported performance is directly listed. Zheng et al.\cite{zheng2020university} was the earliest baseline on this topic with a simple ResNet50 to extract feature. LCM\cite{ding2021practical} solved the lack of satellite images based on location classification, and then LPN\cite{wang2021each} first proposed the square-ring partition strategy. USAM\cite{9779991} inherited the partition strategy and proposed to automatically discover representative key points from feature maps and draw attention to the salient regions. PCL\cite{tian2021uav} proposed perspective projection transformation (PPT) and applied conditional generative adversarial nets (CGAN) to reduce angle differences. All of them are listed for comparison. Furthermore, so far as we know, FSRA\cite{2022A} got the top-1 performance among all the published works. It is a transformer baseline, so it is compared in the table as well. 
	
	\begin{figure*}[ht]
		\centering
		\includegraphics[width=1\linewidth]{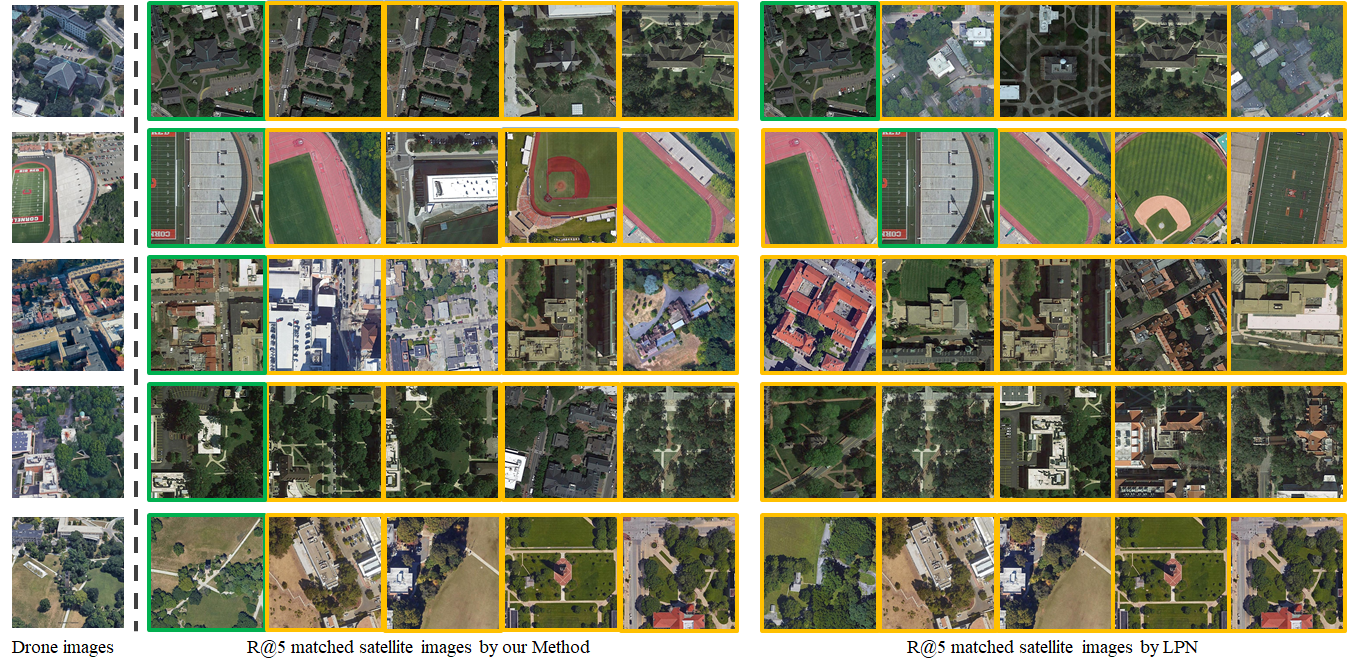}
		\caption{UAV-view geo-localization results from drone to satellite-view images. The left side is the query drone images, the middle lists the top five satellite images matched by our method, and the right side lists the top five matched satellite images using LPN. Especially, the images in the green box are correct matches. There is only one satellite-view image in a group of images with the same localizations.}
		\label{fig:d2s}
	\end{figure*}
	
	\begin{figure*}[h!]
		\centering
		\includegraphics[width=1\linewidth]{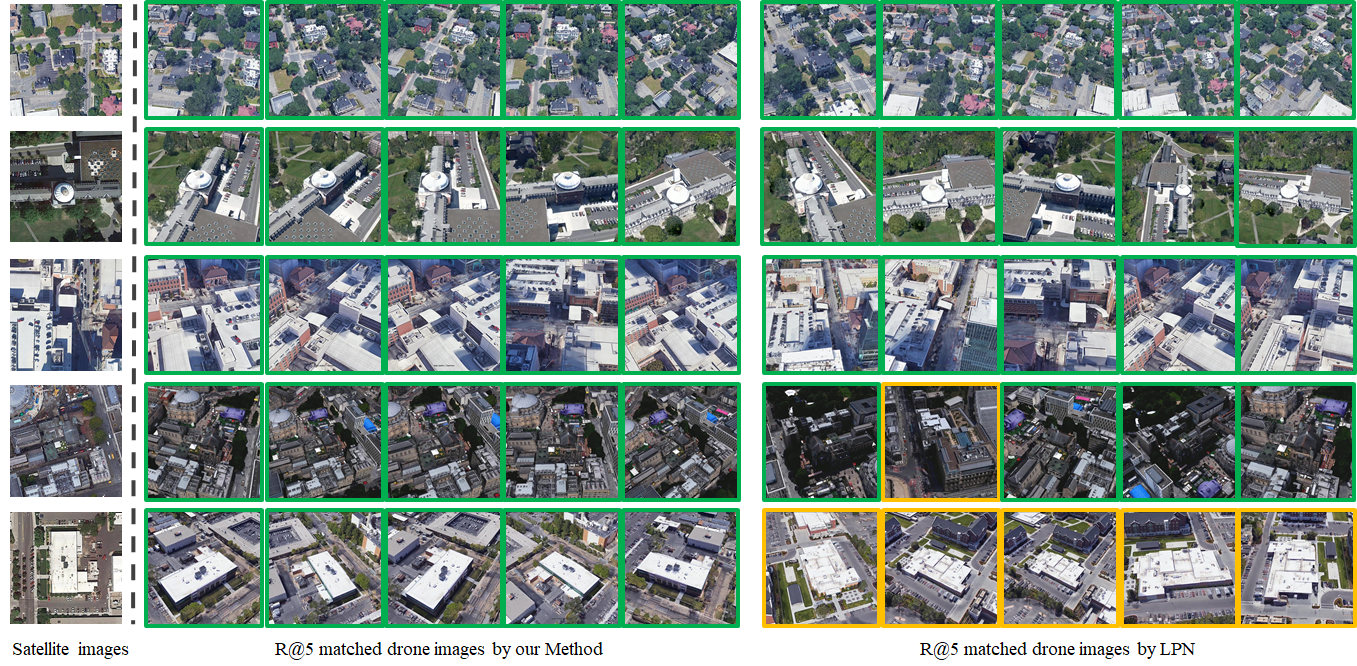}
		\caption{UAV-view geo-localization results from satellite to drone-view images. The left side is the query satellite images, the middle is the top five drone images matched by our method, and the right side lists the top five drone images using LPN. Especially, the images in the green box are correct matches. Commonly there are several drone-view images with the same localizations matching a satellite-view image.}
		\label{fig:std}
	\end{figure*}
	
	Experiments under the settings of geo-localization from drone to satellite view and satellite to drone view are both conducted. Our method achieves the best performance among all the comparisons, which are 84.08\% Recall@1 accuracy and 86.39\% AP from drone to satellite view and 91.44\% Recall@1 accuracy and 82.02\% AP from satellite to drone view. It outperforms LPN\cite{wang2021each} with more than 9\% Recall@1 and 7\% AP improvements, confirming the importance of the dual-square-ring partition. Our performance is superior to FSRA\cite{2022A} as well, thus demonstrating the effectiveness of the proposed style alignment strategy and the HAB attention block.
	
	For further qualitative evaluation, we visualize the retrieval results and show some examples of satellite-matched drone-view images and drone-matched satellite-view images in Fig. 7 and Fig. 8 respectively. The retrieval results of LPN are also listed for comparison. As we can see, our method is qualified for both two tasks. In the task of geo-localization from drone to satellite views, our method successfully selects all the true matched satellite images as Rank@1, which are denoted in the green boxes. But LPN lists the true satellite images as the first and the second Ranks in the top two rows respectively. Especially, when there is no conspicuous architecture in the center of the satellite image, our method would have significant advantages over LPN, as shown from the third to the fifth rows of Fig. 7. It indicates that our dynamic observation module is robust for images with different geographical features. Besides, In the task of geo-localization from satellite to drone, the five recall drone images are all correct by our method. It is not easy for drone images with similar buildings in the center, such as images in the last row. LPN identifies wrong groups of drone images that have similar white roofs as the query satellite image. The probable reason is that the equal weight of different parts in LPN further leads to a false match.

	\subsection{Ablation Studies}
	To investigate the necessity of key components in our model, we design several ablation study experiments on the University-1652 dataset.
	
	{\bf Necessity of SAS}. The style alignment strategy is used for visual style normalization to reduce feature discrepancy in preprocessing. In order to study its effect, we try to add it to LPN\cite{wang2021each} as well. The experimental results on the University-1652 are shown in Table \ref{Tabel2}. There are four groups of results, which are LPN without additional Google training data, LPN without Google data but adding SAS, our method without SAS, and our method. The first two rows of Table \ref{Tabel2} indicate that the SAS helps LPN increase its R@1 accuracy by 4.03\%  and AP by 3.48\% over its original results in the task of geo-localization from drone to satellite. Similar increases also appear in the task of satellite to drone images, they even surpass our performance without the SAS in the third row. however, considerable raises of 6.21\% R@1 and 5.3\% AP are achieved by adding SAS with our method matching from satellite to drone-view images. Stable increases in all the above cases demonstrate the necessity of our style alignment strategy.
	
	\begin{table}[htb]
		\centering
		\caption{Necessity of SAS. Data in \textbf{blod} are the best results, and data with \underline{underlines} are the second best results.}
		\label{Tabel2} 
		\renewcommand\arraystretch{1.5}
		
		\begin{tabular}{c|cc|cc}
			\hline
			\multirow{2}{*}{Method}      & \multicolumn{2}{c|}{Drone-Satellite} & \multicolumn{2}{c}{Satellite-Drone} \\
			& R@1               & AP               & R@1              & AP               \\ \hline
			LPN(w/o Google)              & 74.18             & 77.39            & 85.16            & 73.68            \\
			LPN(w/o Google)+SAS & 78.21    & 80.87  & \underline{89.1}    & \underline{77.21}   \\
			our method(w/o SAS)               & \underline{80.63}             & \underline{83.39}            & 85.23            & 76.72            \\
			\textbf{our method}          & \textbf{84.08}    & \textbf{86.39}   & \textbf{91.44}   & \textbf{82.02}   \\ \hline
		\end{tabular}
	\end{table}
	
	{\bf Effectiveness of HAB}. The hierarchical attention block is an important part of our method, which could acquire accurate feature maps when accompanied by a resnet-based structure. We verify the effectiveness of HAB and the optimal place for HAB in the ResNet-GL structure with a group of experiments. The popular lightweight attention model CBAM is applied for comparison, as shown in Table \ref{Tabel3}. As it is wise not to embed the attention block in the shallow layers of the model due to a lack of semantic information, we consider finding the optimal positional combinations of HABs between Conv3 and Conv5 in ResNet50. Three different positions in the ResNet50-based structure are tested for the attention blocks, which are positions after each layer from conv3 to conv5, denoted as after conv3-5 in the table, positions after layers of conv4 and conv5 respectively, and positions after conv3 and conv5 respectively. Both HAB and CBAM are tested in the above three positions. The comparison results demonstrate that following conv3 and conv5 are the most suitable places for HAB. Moreover, HAB is more effective than CBAM in geo-localization no matter where it is placed.
	
	\begin{table}[]
		\centering
		\caption{Effectiveness of HAB. The best results are in \textbf{bold}, and the second best results are  \underline{underlined}.}
		\label{Tabel3} 
		\renewcommand\arraystretch{1.5}
		\begin{tabular}{c|cc|cc}
			\hline
			\multirow{2}{*}{Method} & \multicolumn{2}{c|}{Drone-\textgreater{}Satellite} & \multicolumn{2}{c}{Satellite-\textgreater{}Drone} \\
			& R@1                      & AP                      & R@1                     & AP                      \\ \hline
			HAB after conv3-5          & 81.03                    & 83.89                   & 85.63                   & 77.12                   \\
			CBAM after conv4-5          & 81.13                    & 83.78                   & 85.65                   & 77.16                   \\
			HAB after conv4-5          & 81.27                    & 84.01                   & 85.69                   & 77.09                   \\
			CBAM after conv3-4         & 81.25                    & 83.93                   & 85.16                   & 73.68                   \\
			HAB after conv3-4          & \underline{82.84}                    & \underline{85.31}                   & \underline{88.45}                   & \underline{78.6}                    \\
			CBAM after conv3,5        & 81.56                    & 84.56                   & 85.49                   & 74.92                   \\
			HAB after conv3,5         & \textbf{84.08}                    & \textbf{86.39}                   & \textbf{91.44}                   & \textbf{82.02}                    \\ \hline
		\end{tabular}
	\end{table}
	
	{\bf Effectiveness of Gem pooling.} Gem pooling is introduced to localize the feature map responses precisely in the situation of massive data including noises. To investigate the effectiveness of Gem pooling, we conduct experiments on our method with several widely used pooling methods, including average pooling, max pooling, and sum pooling. The comparison results are shown in Table \ref{Table_gem}. Max pooling only concerns the maximal response features of each channel. Sum pooling could be considered as a simplified average pooling, which calculates the summation of each channel. The average pooling calculates the average value for each channel. Results in Table \ref{Table_gem} demonstrate that our method with Gem pooling achieves the best performance, which has an improvement of about 2\% than the average pooling. The reason is that it puts more attention on the high response. The average pooling decreases the influence of different feature sizes compared to the sum pooling, but it also dilutes the features, so it achieves the second-best performance. 
	
	\begin{table}[]
		\centering
		\caption{Effectiveness of Gem Pooling. The best results are in \textbf{bold}.}
		\label{Table_gem} 
		\renewcommand\arraystretch{1.5}
		\begin{tabular}{c|clc|cc}
			\hline
			\multirow{2}{*}{Pooling Method} & \multicolumn{3}{c|}{Drone-\textgreater{}Satellite} & \multicolumn{2}{c}{Satellite-\textgreater{}Drone} \\
			& \multicolumn{2}{c}{R@1}       & AP                 & R@1                     & AP                      \\ \hline
			Max pooling                     & \multicolumn{2}{c}{78.12}     & 81.03              & 86.31                   & 75.46                   \\
			Sum pooling                     & \multicolumn{2}{c}{79.89}     & 82.66              & 87.04                   & 77.21                   \\
			Avg pooling                     & \multicolumn{2}{c}{81.29}     & 84.01              & 89.52                   & 80.12                   \\
			Gem pooling                     & \multicolumn{2}{c}{\textbf{84.08}}     & \textbf{86.39}              & \textbf{91.44}                   & \textbf{82.02}                   \\ \hline
		\end{tabular}
	\end{table}
 
    \begin{table}[hb]
		\centering
		
		\caption{Effectiveness of the dc loss. The best results are in \textbf{bold}.}
		\label{tabel_loss} 
		\renewcommand\arraystretch{1.5}
		\begin{tabular}{c|cc|cc}
			\hline
			\multirow{2}{*}{Loss Function} & \multicolumn{2}{c|}{Drone-\textgreater{}Satellite} & \multicolumn{2}{c}{Satellite-\textgreater{}Drone} \\
			& R@1                      & AP                      & R@1                     & AP                      \\ \hline
			Cross-entropy                  & 74.43                    & 77.74                   & 88.69                   & 73.95                   \\
			Cross-entropy+Triplet          & 82.04                    & 84.40                   & 89.23                   & 81.78                   \\
			Cross-entropy+Center           & 82.49                    & 84.93                   & 89.02                   & 81.62                   \\
                Cross-entropy+Triplet+Dc       & 83.63                    & 85.82               & 90.61                   & 81.98                  \\
			Cross-entropy+Center+Dc        & \textbf{84.08}                    & \textbf{86.39}                   & \textbf{91.44}                   & \textbf{82.02}                   \\ \hline
		\end{tabular}
	\end{table}
	{\bf Effectiveness of deconstruction loss}. We propose the deconstruction loss to make full use of the predicted probability of the classifier and push away the cross-correlations of different locations. The combination of the cross-entropy loss, the Deconstruction loss, and the Center loss are utilized in our approach. In this experiment, we try to explore whether deconstruction loss is profitable. As shown in Table \ref{tabel_loss}, we implement the experiments with 4 different loss combinations, including only cross-entropy loss, the cross-entropy loss combined with the triplet loss, the cross-entropy combined with the center loss, the cross-entropy combined with the triplet loss and the deconstruction loss and the cross-entropy loss combined with the center loss and the deconstruction loss. Experiments with only cross-entropy loss show distinct drops on University-1652, which indicates that it is not qualified for geo-localization. The combination of cross-entropy and triplet loss and the combination of cross-entropy and center loss show similar results in the table, but the center loss is capable of gathering inner-class features which is helpful for representative feature extraction in geo-localization. The last two combinations indicate that deconstruction loss is essential for UAV-view geo-localization, while it has further mutually reinforcing effects with class center $c_{n}$ in the center loss. 
	
		\begin{table}[htb]
		\centering
		\caption{Influence of Input Image Size and GFlops of the model with different input image sizes. The best results are in \textbf{bold}.}
		\label{Tabel_size} 
		\renewcommand\arraystretch{1.5}
		\begin{tabular}{c|c|cc|cc}
			\hline
			\multirow{2}{*}{Image size} & \multirow{2}{*}{GFlops} & \multicolumn{2}{c|}{Drone-\textgreater{}Satellite} & \multicolumn{2}{c}{Satellite-\textgreater{}Drone} \\
			&                         & R@1                      & AP                      & R@1                     & AP                      \\ \hline
			256                         & 16.3                    & 84.08                    & 86.39                   & 91.44                   & 82.02                   \\
			384                         & 36.6                    & 85.12                    & 87.42                   & 92.13                   & 83.15                   \\
			512                         & 65.2                    & \textbf{85.79}                    & \textbf{87.96}                   & \textbf{92.52}                   & \textbf{83.39}                   \\
			640                         & 101.8                   & 85.37                    & 87.68                   & 92.26                   & 83.25                   \\ \hline
		\end{tabular}
	\end{table}
 
	{\bf Influence of input image size.} Input images of small size may affect feature extraction on image details, but could effectively economize the training and evaluation cost. Large image size is on the contrary. To investigate the influence of input image size, two groups of experiments are conducted, including geo-localization from drone to satellite views and from satellite to drone views, both of which test four different image sizes as input. The results are shown in Table \ref{Tabel_size}. Besides, GFlops of the model with different input image sizes are listed referring to comparisons on computational costs. The best performance on geo-localization both from drone to satellite-view and satellite to drone-view occurs when the image size is 512. Moreover, when the image size increases from 256 to 512, Rank@1 and AP both have stepwise improvements. But when the image size is bigger than 512, its performance declines. The fluctuation of the results from 256 image size to 640 is under 2\%, which means our method is robust with different image sizes. However, the GFlops of the model with 512px size is quintuple that of the model with 256px size. Therefore, we recommend 256 image size as input because it produces the most efficient results. 
		\begin{table}[htb]
		\centering
		\caption{Influence of center part size.}
		\label{center_size} 
		\renewcommand\arraystretch{1.5}
		\begin{tabular}{c|cc|cc}
			\hline
			\multirow{2}{*}{Ratio of $W_c/W$} & \multicolumn{2}{c|}{Drone-\textgreater{}Satellite} & \multicolumn{2}{c}{Satellite-\textgreater{}Drone} \\
			& R@1                      & AP                      & R@1                     & AP                      \\ \hline
			0.25x                 &   77.64                  & 79.24                   & 86.34                   & 74.57                   \\
			0.75x                 &   81.23                  & 83.40                   & 88.13                   & 81.38                   \\
			0.5x                  & \textbf{84.08}                    & \textbf{86.39}                   & \textbf{91.44}                   & \textbf{82.02}                   
                              \\ \hline
		\end{tabular}
	\end{table}

        \begin{table}[b]
		\centering
		\caption{Influence of rotation on the query drone images.}
		\label{rotation} 
		\renewcommand\arraystretch{1.2}
		\begin{tabular}{c|cc|cc}
			\hline
			\multirow{2}{*}{Rotation Angle} & \multicolumn{2}{c|}{Our Method} & \multicolumn{2}{c}{LPN}\\
			& R@1                      & AP                      & R@1                     & AP                      \\ \hline
	$0^{\circ}$     &   84.08           & 86.39        & 75.93   & 79.14		\\
   $16^{\circ}$         &   83.73          & 85.83         & 75.64   & 78.86  \\
    $45^{\circ}$        &   82.03         &83.74           & 73.04          & 75.62    \\
    $67^{\circ}$         &  81.08         & 82.59            & 70.39          &74.09 \\
    $90^{\circ}$         & 79.02        & 82.14             & 68.80         & 72.67 \\  
    $180^{\circ}$         & 82.78        & 84.23             & 70.76         & 74.47   
    \\
    $270^{\circ}$         & 78.97       & 82.03             & 69.06         & 72.49   
    \\ \hline
		\end{tabular}
	\end{table}
        \begin{table}[b]
		\centering
		\caption{Influence of shift on the query drone images.}
		\label{shift} 
		\renewcommand\arraystretch{1.2}
		\begin{tabular}{c|cc|cc}
			\hline
			\multirow{2}{*}{Shifted Pixels} & \multicolumn{2}{c|}{Our Method} & \multicolumn{2}{c}{LPN}\\
			& R@1                      & AP                      & R@1                     & AP                      \\ \hline
	0     &   84.08                & 86.39                  & 75.93                    & 79.14		\\
        10   &   82.31                 & 84.83                  & 75.26                & 78.57                  \\
			20                  &   80.12                  &81.88                   & 72.37                   & 76.04                   
                              \\
        30                  &   75.09                    &75.93                   &  ---                   & ---                  
                              \\
                              \hline
		\end{tabular}
	\end{table}
    {\bf Influence of center part size.} The central part of our approach is dominant. To investigate the influence of center part size, based on the square-ring partitioning strategy, several center part size settings were compared: 0.75x, 0.5x, and 0.25x, respectively representing the aspect ratio of the center width ($W_c$) and the whole image width ($W$). As shown in Table \ref{center_size}, If the length ratio is 0.25, the performance of our approach is severely degraded. We assumed that information is not sufficient for the hierarchical attention block when the center part is too small. If it is equal to 0.75, the performance also decreases by about 3\%. The reason is assumed to be that due to the poor selectivity of surroundings, the surrounding information in the center partition is discarded. Therefore, the selectivity of the surroundings and the center is balanced when $W_c/W$ is 0.5x. 
        
		\begin{table*}[htb]
		\centering
		\label{tranfer} 
		\caption{Transfer learning from University-1652 to the Sues-200. }
		\renewcommand\arraystretch{1.5}
		\setlength{\tabcolsep}{6mm}
		\begin{tabular}{ccccccccc}
			\hline
			\multicolumn{9}{c}{Sues-200}                                                  \\ 
			\multicolumn{9}{c}{Drone-\textgreater{}Satellite(Training on University-1652)}                                                                                                         \\ \hline
			\multicolumn{1}{c|}{\multirow{2}{*}{method}} & \multicolumn{2}{c|}{150m}          & \multicolumn{2}{c|}{200m}          & \multicolumn{2}{c|}{250m}          & \multicolumn{2}{c}{300m} \\
			\multicolumn{1}{c|}{}                        & R@1   & \multicolumn{1}{c|}{AP}    & R@1   & \multicolumn{1}{c|}{AP}    & R@1   & \multicolumn{1}{c|}{AP}    & R@1         & AP         \\ \hline
			\multicolumn{1}{c|}{SE-ResNet\cite{2022arXiv220410704Z}}       & 32.33 & \multicolumn{1}{c|}{39.01} & 40.55 & \multicolumn{1}{c|}{47.25} & 45.63 & \multicolumn{1}{c|}{52.27} & 50.05       & 56.26      \\ \hline
			\multicolumn{1}{c|}{LPN\cite{wang2021each}}       & 37.47 & \multicolumn{1}{c|}{43.51} & 45.03 & \multicolumn{1}{c|}{51.67} & 50.83 & \multicolumn{1}{c|}{57.42} & 53.30       & 59.65      \\ \hline
			\multicolumn{1}{c|}{Ours(w/o SAS)}           & 45.67 & \multicolumn{1}{c|}{50.95} & 56.48 & \multicolumn{1}{c|}{61.50}  & 61.87 & \multicolumn{1}{c|}{66.49} & 64.65       & 68.92      \\
			\multicolumn{1}{c|}{\textbf{Ours}}           & \textbf{47.65} & \multicolumn{1}{c|}{\textbf{53.31}} & \textbf{60.35} & \multicolumn{1}{c|}{\textbf{64.6}} & \textbf{67.55} & \multicolumn{1}{c|}{\textbf{71.47}} & \textbf{73.05} & \textbf{76.39} \\ \hline
		\end{tabular}
	\end{table*}

    {\bf Influence of shift and rotation.}
    Drone view images inevitably suffer from rotation, perspective shift, and bounding box offset in the upstream detection task. These conditions may make the central salient feature shift and split by the dual square-ring strategy. To explore whether our method can cope with rotations and shifts, we carry out two experiments and show their results in Table \ref{rotation} and Table \ref{shift} respectively. First, we rotate the query drone images from 0 degrees to 270 degrees and compare our results with the LPN's \cite{wang2021each}. As shown in Table \ref{rotation}, both our method and the LPN have a slight decrease in performance as the rotation angle increases from vertical to horizontal. When the rotation angle reaches 90 and 270 degrees, the LPN drops by about 7\%, while our AP only drops by 4\%. Second, we shifted the image from 10 pixels to 30 pixels, the results are shown in Table \ref{shift}, both our method and the LPN exhibit a partial decrease as the number of shifted pixels increases, and our method incurs a relatively significant decrease in performance at a 30-pixel shifted. We assume that it is because when the landmark shifts more than 20 pixels, surrounding noise would hamper the assessment of the landmark in HAB. On the other hand, the other channel in HAB, which inputs the whole image, is affected by the black padding as well. But in terms of overall performance, our method have the ability to better overcome the influence of shifts and rotations.

    {\bf Influence of drone flight altitudes.} The scope of satellite-view images in dataset University-1652 is fixed, but the scope of the drone-view images varies following the variations of the drone altitudes. Therefore, we would like to show experimental matching results on geo-localization images from drone to satellite with different drone flight altitudes. The drone images are divided into three altitudes as queries: long, middle, and short in the same way as LPN\cite{wang2021each} and FSRA\cite{2022A}. We compare our model with these two state-of-the-art methods in the experiment. The performance is reported in Fig. 9. Our method achieves the best AP and Rank@1 when a drone is flying at short and middle altitudes, as well as the average performance of all altitudes. We achieve the second-best performance under the condition of long altitudes. The possible reason is that the targets in the drone images with long altitudes are much smaller than in the other two situations, which is in favor of contextual segmentation-based FSRA. But our method has a large improvement of about 15\% in both Rank@1 and AP compared to LPN in experiments of long altitude.
 
	\begin{figure}[htb]
		\centering  
		\subfloat[AP with different drone altitudes.]{   
			\begin{minipage}{.47\linewidth}
                \captionsetup{font=small}
				\centering    
				\includegraphics[scale=0.3]{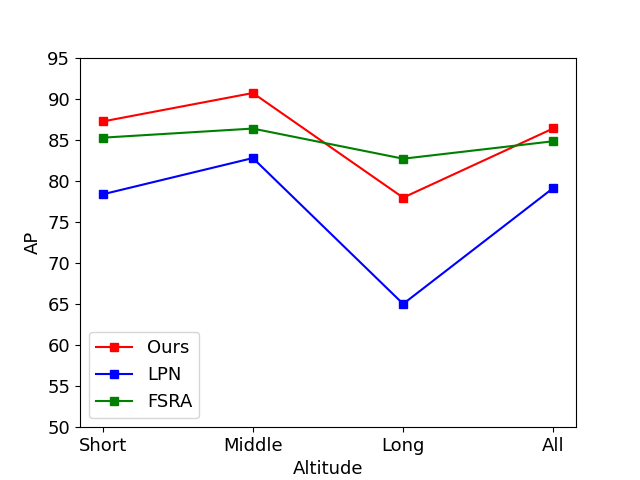}  
			\end{minipage}
		}
		\hfill
		\subfloat[R@1 with different drone altitudes.]{ 
			\begin{minipage}{.47\linewidth}
				\centering    
				\includegraphics[scale=0.3]{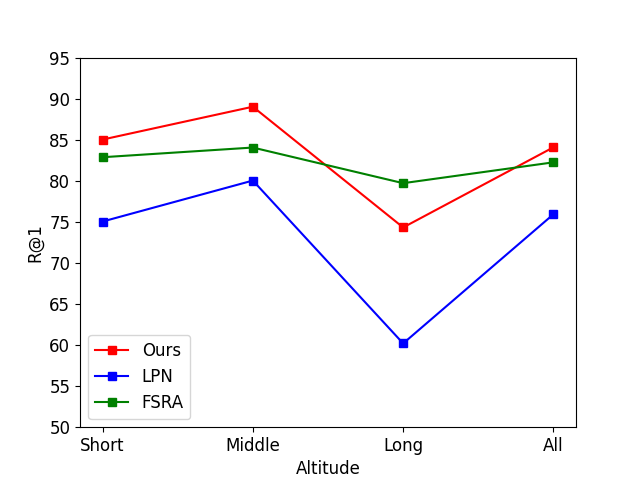}
			\end{minipage}
		}
		\hfill
		\caption{Influence of the drone altitude, where the R@1 evaluation metric results are shown in (a) and the results of the AP evaluation metric are shown in (b).}    
		\label{fig:1}    
	\end{figure}

	\subsection{Transfer Learning}
	The purpose of transfer learning is to verify the robustness and effectiveness of our method in cross-domain UAV-view geo-localization. We implement the experiments by transferring the model trained on the University-1652 to the Sues-200 to evaluate its UAV-view geo-localization ability. In preprocessing, the style alignment strategy is employed on the Sues-200 dataset. Especially, our SAS benefits from rapid visual feature alignment between drone and satellite-view images in different datasets, and it is skilled in contextual analysis. The performance of our model is shown in Table X with multiple altitudes, including 150m, 200m, 250m, and 300m. SE-ResNet, LPN, and our method (with or without SAS) are listed for comparison. Our method achieves the best performance for all altitudes on the Sues-200 with the model trained on the University-1652. Compared to LPN, we have about 10\% average improvement in Rank@1 and AP even without SAS, which means our HAB blocks are robust in cross-domain feature alignment. Then after adding SAS, our performance further improves more than 7\%, 5\%, 3\%, and 2\% in experiments with 300m, 250m, 200m, and 150m flight altitudes respectively. It demonstrates that SAS is an effective way to reduce the domain gap and is also effective in transfer learning.

	\subsection{Discussions on the style alignment}
\begin{figure}[t]
		\centering
		\includegraphics[width=0.9\linewidth]{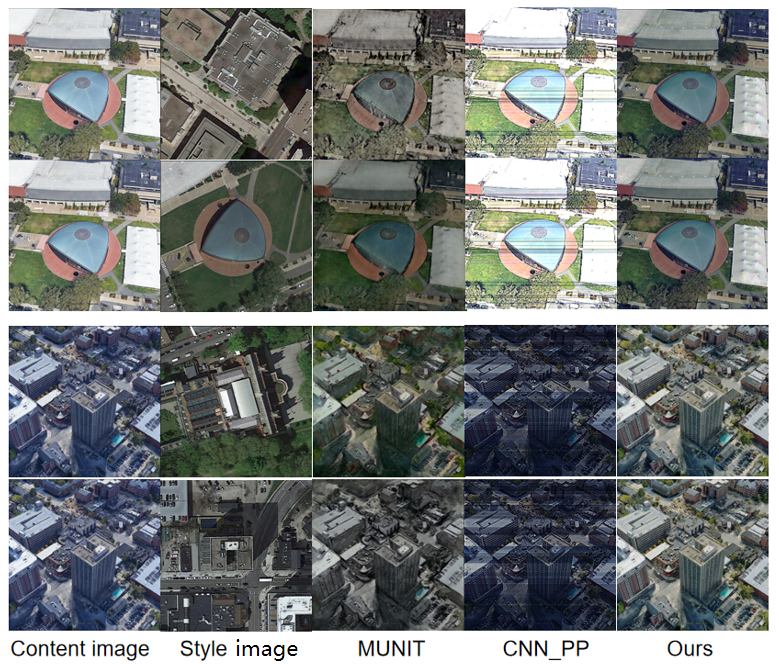}
		\caption{Visualization comparison of several style alignment methods. There are two samples of drone view images. The first column is the original image, and the second column is the style reference. The rest columns present the style transfer results of MUNIT, CNN\_PP, and ours respectively.}
		\label{fig:style}
	\end{figure}

	Our purpose is to adjust the light conditions and the camera styles of the drone view to be consistent with the satellite view. We display several methods that may accomplish the above purposes, as shown in Fig. \ref{fig:style}. Regarding the common methods of image style transfer based on generative adversarial networks, image adaptation, and traditional image processing, we selected one representative method for each category and presented their visual comparison.
    
    It should be noted that we do not have the drone image in standard light conditions as the ground truth. So the supervised methods like CycleGAN\cite{Zhu2017UnpairedIT}, Pix2Pix\cite{Isola2016ImagetoImageTW} are not applicable. MUNIT\cite{Huang2018MultimodalUI} is a GAN-based method that could divide the image into a content part and a style part, so we conduct some experiments by the MUNIT. As we can see in Fig. \ref{fig:style}, the MUNIT-based method requires a reference image for style transfer. However, it is not possible to determine which satellite image is the style reference for the drone image during testing. Different satellite images will lead to different transformation results for the drone images. The first two rows and the bottom two rows of Fig. \ref{fig:style} show two sets of drone images, respectively, and they get completely different conversion results after style transfer by referring them to different satellite images respectively. Results are shown in the third column. In addition, MUNIT cannot perfectly separate style and content either. It is obvious from the images in the third column that the texture and details of the images have been changed by the corresponding reference images. 
    
    On the other hand, IA-Seg\cite{Liu2022ImprovingND} uses a CNN model named CNN\_PP to predict image filter parameters which adaptively adjust the exposure, gamma, contrast, and sharpening of the image for visual style transfer. It is a representative image adaptation method. Its conversion results are shown in column 4. We assume that IA-Seg needs pixel-level supervision for filter parameter prediction, but we only have category-level supervision. So the style conversion results are poor by IA-Seg in our experiments. As we can see images in the last column, the style transfer results are more stable than all the other two kinds of methods. It is also fast and needs no training.
	\subsection{Discussions on model size}
\begin{figure}[htb]
		\centering
		\includegraphics[width=0.9\linewidth]{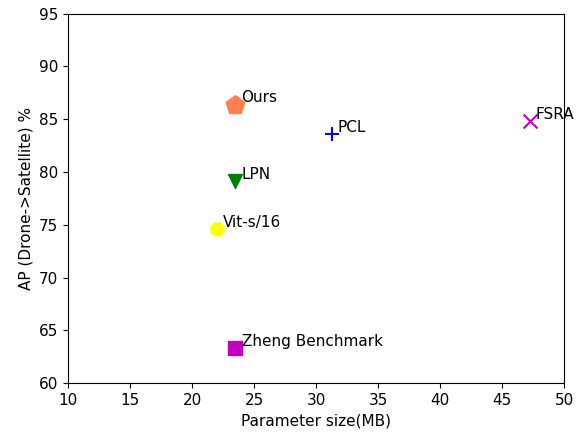}
		\caption{Comparisons of parameter sizes and AP values from drone to satellite view with state-of-the-art methods.}
		\label{fig.10}
	\end{figure}

	To demonstrate our model is effective and light-weighted, we compare the model size and AP with the backbones of existing works without classifier module in this section. As shown in Fig. \ref{fig.10}, the abscissa is the parameter size, which is measured by MB. The ordinates are AP values. LPN\cite{wang2021each}, Vit-s\cite{2022A}, Zheng\cite{zheng2020university}, PCL\cite{tian2021uav}, FSRA\cite{2022A} and our method are plotted with different shape and color. The parameter size of our method is only 23.6M, which is similar to LPN, Vit-s, and Zheng, but our AP value is much higher than these three methods. FSRA has the top results among all the published works by now, but its model size is about twice ours and its AP is lower. Specifically, on the ResNet50 backbone, our hierarchical attention blocks add only 0.02GFlops.
        
	\subsection{Visualization}
\begin{figure*}[t]
		\centering  
		\subfloat[]{   
			\begin{minipage}{.22\linewidth}
				\includegraphics[scale=0.3]{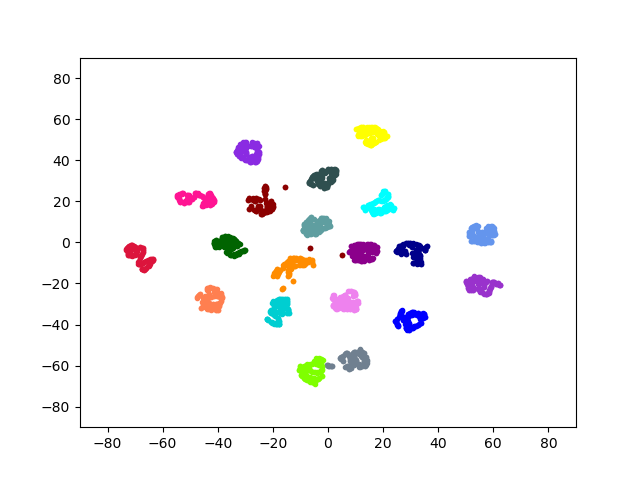}  
			\end{minipage}
		}
		\hfill
		\subfloat[]{ 
			\begin{minipage}{.22\linewidth}
				\includegraphics[scale=0.3]{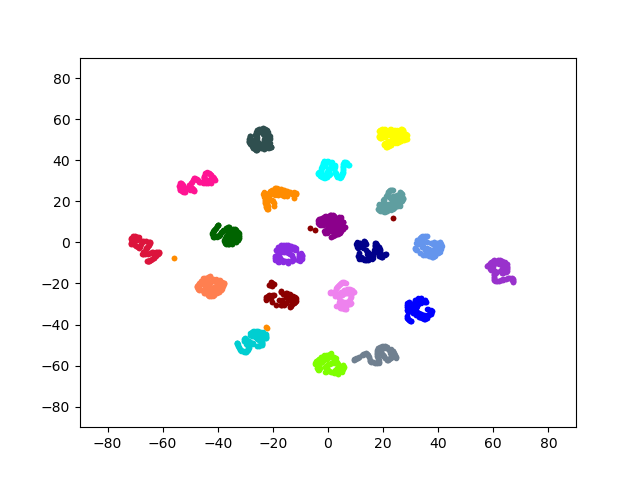}
			\end{minipage}
		}
		\hfill
		\subfloat[]{ 
			\begin{minipage}{.22\linewidth}
				\includegraphics[scale=0.3]{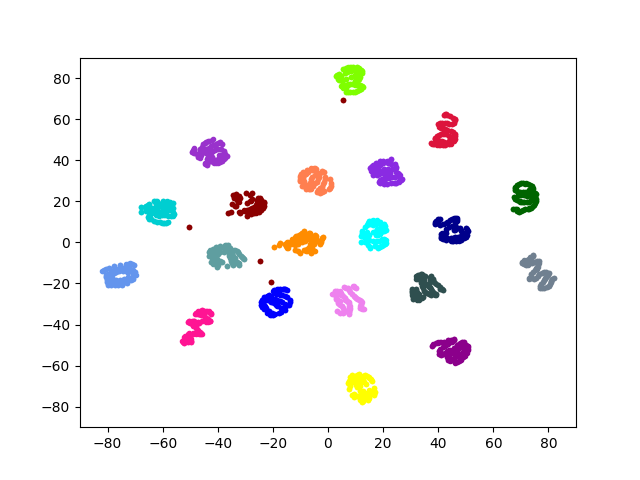}
			\end{minipage}
		}
		\hfill
		\caption{Visualization of feature distributions. (a) and (b) show the feature distributions of ResNet50 and the LPN, respectively. For each image, there are 20 randomly selected classes. (c) shows the feature distribution of our method with Dc Loss.}    
		\label{fig:feature}    
	\end{figure*}

	In this section, visualizations of feature distributions are shown in Fig. \ref{fig:feature}. There are three images (a), (b), and (c), showing feature distributions of 20 randomly selected classes generated by ResNet50, LPN, and our method with Dc Loss, respectively. As shown in Fig. \ref{fig:feature} (a) and (b), feature clusters are closely gathered. In contrast, because the Dc loss pushes the decision boundaries apart and wraps each cluster tightly, the distances between feature clusters are more dispersed.
	
\begin{figure*}[ht]
		\centering
		\includegraphics[width=1\linewidth]{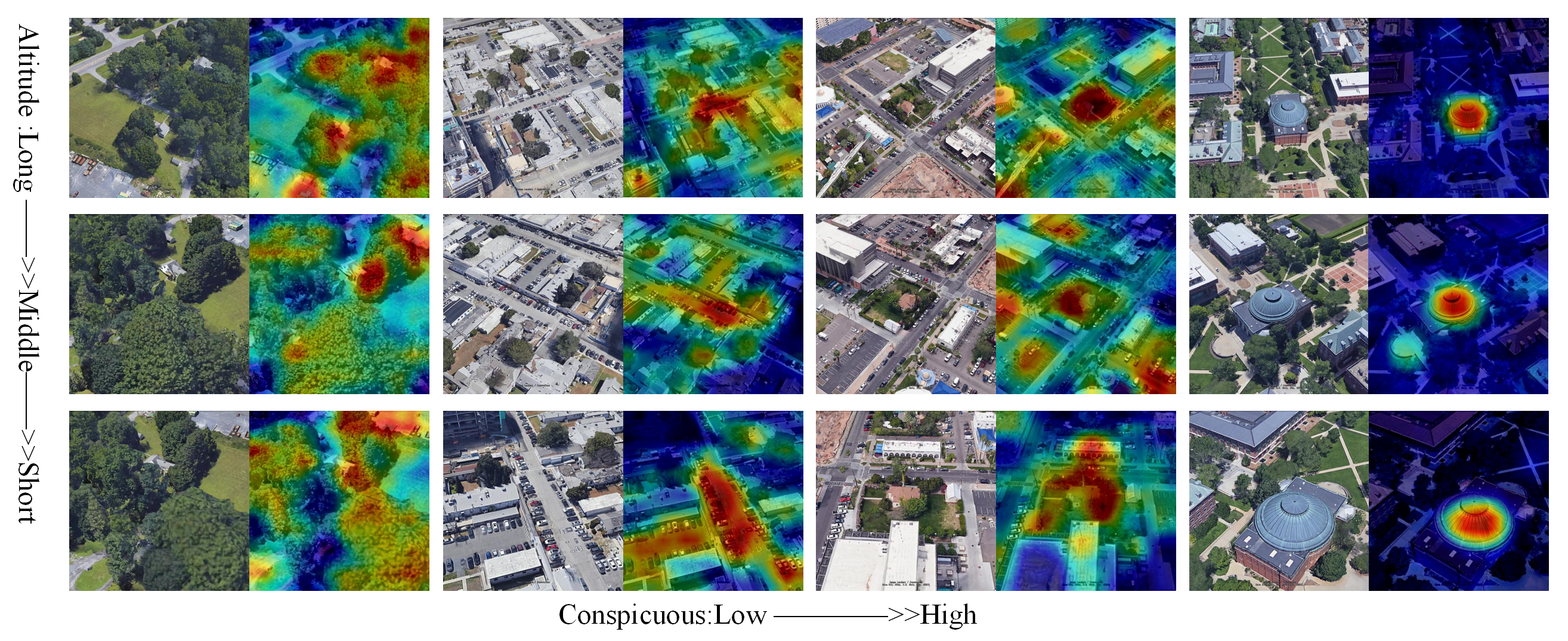}
		\caption{Visualization of heatmaps of our method. There are four groups of drone view images and the corresponding heatmaps generated by GradCam.From left to right, the center architectures have gradually conspicuous features. And from top to down, the altitudes of the drone are gradually low.}
		\label{fig.11}
	\end{figure*}

	Then we randomly show several drone-view images from the test dataset with different altitudes and architecture conspicuousness, and then visualize their heat maps generated by GradCam\cite{2016arXiv161002391S}. In Fig. \ref{fig.11}, there are 12 groups of drone view images and their corresponding heatmaps of our method. The architectural conspicuousness in the center part of images gradually increases from left to right. Correspondingly, the heatmaps from left to right demonstrate our dynamic observation method could focus on distinct information in the image. When there is no well-marked building in the center, the surroundings would attract considerable attention, as images shown in the first column. However, if the center has distinct features in the image, our attention would be focused on prominent architectures, as shown in the last column. In addition, as the drone flight altitudes gradually decrease from the top row to the bottom rows, it shows that the responses of different altitudes are consistent which means that our dynamic observation method has a good performance at different altitudes.  
	
	\section{Conclusion}
	In this paper, we propose a style alignment based dynamic observation method for UAV-View Geo-Localization, which contains three parts. First, the style alignment strategy is proposed to transfer various visual styles of drone-view images to satellite views in a fast and effective way. Then, the dynamic observation module extract features in a human observation manner, which dynamically considers the significance of the center building and the surrounding. Finally, the Deconstruction loss is applied to compact the feature of every geo-tag and offset the flaw of the Cross-entropy loss. Our method achieves superior performance in benchmark datasets. 
 
	As an extension of our research, we are in the process of studying the self-supervised depth estimation in drone viewing using continuous frames provided by the dataset. It helps us to understand the images in higher dimensions and improves the performance of the localization. Besides, as the remote sensing data usually tend to suffer from various degradation, noise effects, or variabilities in the process of imaging\cite{8528557}, we plan to try cross-view geo-localization of degraded satellite images in the future.
	
	\section*{Acknowledgments}
	This work is supported by the Local College Capacity Building Project of Shanghai Municipal Science and Technology Commission under Grant 20020500700, and by the National Natural Science Foundation of China(No.61802250)

	
		\bibliographystyle{IEEEtran}
		\bibliography{ref}

\begin{thebibliography}{10}
\providecommand{\url}[1]{#1}
\csname url@samestyle\endcsname
\providecommand{\newblock}{\relax}
\providecommand{\bibinfo}[2]{#2}
\providecommand{\BIBentrySTDinterwordspacing}{\spaceskip=0pt\relax}
\providecommand{\BIBentryALTinterwordstretchfactor}{4}
\providecommand{\BIBentryALTinterwordspacing}{\spaceskip=\fontdimen2\font plus
\BIBentryALTinterwordstretchfactor\fontdimen3\font minus \fontdimen4\font\relax}
\providecommand{\BIBforeignlanguage}[2]{{%
\expandafter\ifx\csname l@#1\endcsname\relax
\typeout{** WARNING: IEEEtran.bst: No hyphenation pattern has been}%
\typeout{** loaded for the language `#1'. Using the pattern for}%
\typeout{** the default language instead.}%
\else
\language=\csname l@#1\endcsname
\fi
#2}}
\providecommand{\BIBdecl}{\relax}
\BIBdecl

\bibitem{2020Where}
Y.~Shi, X.~Yu, D.~Campbell, and H.~Li, ``Where am i looking at? joint location and orientation estimation by cross-view matching,'' in \emph{2020 IEEE/CVF Conference on Computer Vision and Pattern Recognition (CVPR)}, 2020, pp. 4063--4071.

\bibitem{liu2019lending}
L.~Liu and H.~Li, ``Lending orientation to neural networks for cross-view geo-localization,'' in \emph{Proceedings of the IEEE/CVF Conference on Computer Vision and Pattern Recognition}, 2019, pp. 5624--5633.

\bibitem{9684950}
Z.~Zeng, Z.~Wang, F.~Yang, and S.~Satoh, ``Geo-localization via ground-to-satellite cross-view image retrieval,'' \emph{IEEE Transactions on Multimedia}, pp. 1--1, 2022.

\bibitem{2022arXiv220701768W}
X.~{Wang}, D.~{Zeng}, Q.~{Zhao}, and S.~{Li}, ``{Rank-Based Filter Pruning for Real-Time UAV Tracking},'' \emph{arXiv e-prints}, p. arXiv:2207.01768, Jul. 2022.

\bibitem{2022arXiv220611499J}
S.~{Jiang}, Q.~{Li}, W.~{Jiang}, and W.~{Chen}, ``{Parallel Structure from Motion for UAV Images via Weighted Connected Dominating Set},'' \emph{arXiv e-prints}, p. arXiv:2206.11499, Jun. 2022.

\bibitem{9779991}
J.~Lin, Z.~Zheng, Z.~Zhong, Z.~Luo, S.~Li, Y.~Yang, and N.~Sebe, ``Joint representation learning and keypoint detection for cross-view geo-localization,'' \emph{IEEE Transactions on Image Processing}, vol.~31, pp. 3780--3792, 2022.

\bibitem{wang2021each}
T.~Wang, Z.~Zheng, C.~Yan, J.~Zhang, Y.~Sun, B.~Zhenga, and Y.~Yang, ``Each part matters: Local patterns facilitate cross-view geo-localization,'' \emph{IEEE Transactions on Circuits and Systems for Video Technology}, 2021.

\bibitem{tian2021uav}
X.~Tian, J.~Shao, D.~Ouyang, and H.~T. Shen, ``Uav-satellite view synthesis for cross-view geo-localization,'' \emph{IEEE Transactions on Circuits and Systems for Video Technology}, 2021.

\bibitem{Zhu2017UnpairedIT}
J.-Y. Zhu, T.~Park, P.~Isola, and A.~A. Efros, ``Unpaired image-to-image translation using cycle-consistent adversarial networks,'' \emph{2017 IEEE International Conference on Computer Vision (ICCV)}, pp. 2242--2251, 2017.

\bibitem{Isola2016ImagetoImageTW}
P.~Isola, J.-Y. Zhu, T.~Zhou, and A.~A. Efros, ``Image-to-image translation with conditional adversarial networks,'' \emph{2017 IEEE Conference on Computer Vision and Pattern Recognition (CVPR)}, pp. 5967--5976, 2016.

\bibitem{2022A}
M.~Dai, J.~Hu, J.~Zhuang, and E.~Zheng, ``A transformer-based feature segmentation and region alignment method for uav-view geo-localization,'' \emph{IEEE Transactions on Circuits and Systems for Video Technology}, vol.~32, no.~7, pp. 4376--4389, 2022.

\bibitem{2022arXiv220410704Z}
R.~{Zhu}, ``{SUES-200: A Multi-height Multi-scene Cross-view Image Benchmark Across Drone and Satellite},'' \emph{arXiv e-prints}, p. arXiv:2204.10704, Apr. 2022.

\bibitem{2022Vision}
M.~Dai, J.~Huang, J.~Zhuang, W.~Lan, Y.~Cai, and E.~Zheng, ``Vision-based uav localization system in denial environments,'' \emph{arXiv e-prints}, 2022.

\bibitem{9598903}
X.~Wu, D.~Hong, and J.~Chanussot, ``Convolutional neural networks for multimodal remote sensing data classification,'' \emph{IEEE Transactions on Geoscience and Remote Sensing}, vol.~60, pp. 1--10, 2022.

\bibitem{zheng2020university}
Z.~Zheng, Y.~Wei, and Y.~Yang, ``University-1652: A multi-view multi-source benchmark for drone-based geo-localization,'' in \emph{Proceedings of the 28th ACM international conference on Multimedia}, 2020, pp. 1395--1403.

\bibitem{2017Beyond}
Y.~Sun, L.~Zheng, Y.~Yang, Q.~Tian, and S.~Wang, ``Beyond part models: Person retrieval with refined part pooling (and a strong convolutional baseline),'' in \emph{European Conference on Computer Vision}, 2017.

\bibitem{2017arXiv171108184Z}
X.~{Zhang}, H.~{Luo}, X.~{Fan}, W.~{Xiang}, Y.~{Sun}, Q.~{Xiao}, W.~{Jiang}, C.~{Zhang}, and J.~{Sun}, ``{AlignedReID: Surpassing Human-Level Performance in Person Re-Identification},'' \emph{arXiv e-prints}, p. arXiv:1711.08184, Nov. 2017.

\bibitem{2022Multiple}
T.~Wang, Z.~Zheng, Y.~Sun, T.~S. Chua, Y.~Yang, and C.~Yan, ``Multiple-environment self-adaptive network for aerial-view geo-localization,'' 2022.

\bibitem{ding2021practical}
L.~Ding, J.~Zhou, L.~Meng, and Z.~Long, ``A practical cross-view image matching method between uav and satellite for uav-based geo-localization,'' \emph{Remote Sensing}, vol.~13, no.~1, p.~47, 2021.

\bibitem{2017A}
F.~K.~V. Evert, G.~W. A.~M. van~der Heijden, L.~A.~P. Lotz, G.~Polder, A.~Lamaker, A.~D. Jong, M.~C. Kuyper, E.~J.~K. Groendijk, and N.~T. V.~D. Zalm, ``A mobile field robot with vision-based detection of volunteer potato plants in a corn crop,'' \emph{Weed Technology}, vol.~20, no.~4, pp. 853--861, 2017.

\bibitem{2015arXiv150303832S}
F.~Schroff, D.~Kalenichenko, and J.~Philbin, ``Facenet: A unified embedding for face recognition and clustering,'' in \emph{2015 IEEE Conference on Computer Vision and Pattern Recognition (CVPR)}, 2015, pp. 815--823.

\bibitem{2006Dimensionality}
R.~Hadsell, S.~Chopra, and Y.~Lecun, ``Dimensionality reduction by learning an invariant mapping,'' in \emph{2006 IEEE Computer Society Conference on Computer Vision and Pattern Recognition (CVPR'06)}, 2006.

\bibitem{2022Hierarchical}
J.~Shao and X.~Ma, ``Hierarchical pseudo-label learning for one-shot person re-identification,'' \emph{Applied Intelligence}, vol.~52, no.~8, pp. 9225--9238, 2022.

\bibitem{2020Dual}
Z.~Zheng, L.~Zheng, M.~Garrett, Y.~Yang, M.~Xu, and Y.~Shen, ``Dual-path convolutional image-text embeddings with instance loss,'' \emph{ACM Transactions on Multimedia Computing, Communications, and Applications (TOMM)}, 2020.

\bibitem{hu2018cvm}
S.~Hu, M.~Feng, R.~M. Nguyen, and G.~H. Lee, ``Cvm-net: Cross-view matching network for image-based ground-to-aerial geo-localization,'' in \emph{Proceedings of the IEEE Conference on Computer Vision and Pattern Recognition}, 2018, pp. 7258--7267.

\bibitem{2020A}
H.~Luo, W.~Jiang, Y.~Gu, F.~Liu, X.~Liao, S.~Lai, and J.~Gu, ``A strong baseline and batch normalization neck for deep person re-identification,'' \emph{IEEE}, no.~10, 2020.

\bibitem{hu2018squeeze}
J.~Hu, L.~Shen, and G.~Sun, ``Squeeze-and-excitation networks,'' in \emph{Proceedings of the IEEE conference on computer vision and pattern recognition}, 2018, pp. 7132--7141.

\bibitem{woo2018cbam}
S.~Woo, J.~Park, J.-Y. Lee, and I.~S. Kweon, ``Cbam: Convolutional block attention module,'' in \emph{Proceedings of the European conference on computer vision (ECCV)}, 2018, pp. 3--19.

\bibitem{rs11080917}
\BIBentryALTinterwordspacing
X.~Pan, F.~Yang, L.~Gao, Z.~Chen, B.~Zhang, H.~Fan, and J.~Ren, ``Building extraction from high-resolution aerial imagery using a generative adversarial network with spatial and channel attention mechanisms,'' \emph{Remote Sensing}, vol.~11, no.~8, 2019. [Online]. Available: \url{https://www.mdpi.com/2072-4292/11/8/917}
\BIBentrySTDinterwordspacing

\bibitem{GraphCN}
D.~Hong, L.~Gao, J.~Yao, B.~Zhang, A.~J. Plaza, and J.~Chanussot, ``Graph convolutional networks for hyperspectral image classification,'' \emph{IEEE Transactions on Geoscience and Remote Sensing}, vol.~59, pp. 5966--5978, 2020.

\bibitem{9386248}
H.~Chen, W.~Li, and Z.~Shi, ``Adversarial instance augmentation for building change detection in remote sensing images,'' \emph{IEEE Transactions on Geoscience and Remote Sensing}, vol.~60, pp. 1--16, 2022.

\bibitem{8382272}
F.~Radenović, G.~Tolias, and O.~Chum, ``Fine-tuning cnn image retrieval with no human annotation,'' \emph{IEEE Transactions on Pattern Analysis and Machine Intelligence}, vol.~41, no.~7, pp. 1655--1668, 2019.

\bibitem{arxiv20reidsurvey}
M.~Ye, J.~Shen, G.~Lin, T.~Xiang, L.~Shao, and S.~C.~H. Hoi, ``Deep learning for person re-identification: A survey and outlook,'' \emph{arXiv preprint arXiv:2001.04193}, 2020.

\bibitem{deng2022insclr}
Z.~Deng, Y.~Zhong, S.~Guo, and W.~Huang, ``Insclr: Improving instance retrieval with self-supervision,'' in \emph{Proceedings of the AAAI Conference on Artificial Intelligence}, vol.~36, no.~1, 2022, pp. 516--524.

\bibitem{zhuang2021faster}
J.~Zhuang, M.~Dai, X.~Chen, and E.~Zheng, ``A faster and more effective cross-view matching method of uav and satellite images for uav geolocalization,'' \emph{Remote Sensing}, vol.~13, no.~19, p. 3979, 2021.

\bibitem{2016A}
Y.~Wen, K.~Zhang, Z.~Li, and Y.~Qiao, ``A discriminative feature learning approach for deep face recognition,'' 2016.

\bibitem{lu2022content}
Z.~Lu, T.~Pu, T.~Chen, and L.~Lin, ``Content-aware hierarchical representation selection for cross-view geo-localization,'' in \emph{Proceedings of the Asian Conference on Computer Vision}, 2022, pp. 4211--4224.

\bibitem{Huang2018MultimodalUI}
X.~Huang, M.-Y. Liu, S.~J. Belongie, and J.~Kautz, ``Multimodal unsupervised image-to-image translation,'' in \emph{European Conference on Computer Vision}, 2018.

\bibitem{Liu2022ImprovingND}
W.~Liu, W.~Li, J.~Zhu, M.~Cui, X.~Xie, and L.~Zhang, ``Improving nighttime driving-scene segmentation via dual image-adaptive learnable filters,'' \emph{ArXiv}, vol. abs/2207.01331, 2022.

\bibitem{2016arXiv161002391S}
R.~R. Selvaraju, M.~Cogswell, A.~Das, R.~Vedantam, D.~Parikh, and D.~Batra, ``Grad-cam: Visual explanations from deep networks via gradient-based localization,'' in \emph{2017 IEEE International Conference on Computer Vision (ICCV)}, 2017, pp. 618--626.

\bibitem{8528557}
D.~Hong, N.~Yokoya, J.~Chanussot, and X.~X. Zhu, ``An augmented linear mixing model to address spectral variability for hyperspectral unmixing,'' \emph{IEEE Transactions on Image Processing}, vol.~28, no.~4, pp. 1923--1938, 2019.

\end{thebibliography}

	\vfill
	
\end{document}